\documentclass[runningheads]{llncs}

\usepackage{eccv}

\usepackage{eccvabbrv}

\usepackage{graphicx}
\usepackage{booktabs}

\usepackage[accsupp]{axessibility}  

\usepackage{hyperref}

\usepackage{orcidlink}

\usepackage{multirow}
\usepackage{array}
\renewcommand\arraystretch{1.3} 
\usepackage{makecell}
\usepackage{booktabs}
\usepackage{hhline}
\usepackage{xcolor}
\usepackage{colortbl}

\colorlet{tableheadcolor}{gray!25} 
\colorlet{tablerowcolor}{gray!15} 

\newcommand{\rowcol}{\rowcolor{tablerowcolor}}

\usepackage{bbding}
\newcommand{\PreserveBackslash}[1]{\let\temp=\\#1\let\\=\temp}
\newcolumntype{C}[1]{>{\PreserveBackslash\centering}p{#1}}
\newcolumntype{R}[1]{>{\PreserveBackslash\raggedleft}p{#1}}
\newcolumntype{L}[1]{>{\PreserveBackslash\raggedright}p{#1}}

\newcommand{\modelname}{UMG-CLIP\xspace}
\newcommand{\dataname}{UMG-41M\xspace}

\begin{document}

\title{UMG-CLIP: A Unified Multi-Granularity Vision Generalist for Open-World Understanding} 

\titlerunning{UMG-CLIP for Open-World Understanding}

\author{Bowen Shi\inst{1}$^{\dagger}$, Peisen Zhao\inst{2}$^{\dagger}$,  Zichen Wang\inst{2}, Yuhang Zhang\inst{2},\\ Yaoming Wang\inst{1}, Jin Li\inst{1}, Wenrui Dai\inst{1}, Junni Zou\inst{1},\\ Hongkai Xiong\inst{1}, Qi Tian\inst{2}, and Xiaopeng Zhang\inst{2}\thanks{ Corresponding author. $^{\dagger}$ Equal contribution.}}

\authorrunning{B. Shi et al.}

\institute{Shanghai Jiao Tong University \\
\email{\{sjtu\_shibowen, wang\_yaoming, deserve\_lj, daiwenrui, zoujunni, xionghongkai\}@sjtu.edu.cn}\\
\and
 Huawei Inc.\\
\email{\{pszhao93, zcwang0118, cupcake3419, zxphistory\}@gmail.com},
\email{tian.qi1@huawei.com}}

\maketitle

\begin{abstract}
Vision-language foundation models, represented by Contras-tive Language-Image Pre-training (CLIP), have gained increasing attention for jointly understanding both vision and textual tasks. However, existing approaches primarily focus on training models to match global image representations with textual descriptions, thereby overlooking the critical alignment between local regions and corresponding text tokens. This paper extends CLIP with multi-granularity alignment. Notably, we deliberately construct a new dataset comprising pseudo annotations at various levels of granularities, encompassing image-level, region-level as well as pixel-level captions and tags. Accordingly, we develop a Unified Multi-Granularity learning framework, termed UMG-CLIP, which simultaneously empowers the model with versatile perception abilities across different levels of detail. With parameter efficient tuning, UMG-CLIP surpasses current widely used CLIP variants and achieves state-of-the-art performance on diverse image understanding benchmarks, including open-world recognition, retrieval, semantic segmentation, and panoptic segmentation tasks. We believe that UMG-CLIP represents a valuable advancement in vision-language foundation models. The code is available at \url{https://github.com/lygsbw/UMG-CLIP}.
\end{abstract}    
\section{Introduction}
\label{sec:intro}

\begin{figure}[t]
    \centering
    \includegraphics[width=0.8\linewidth]{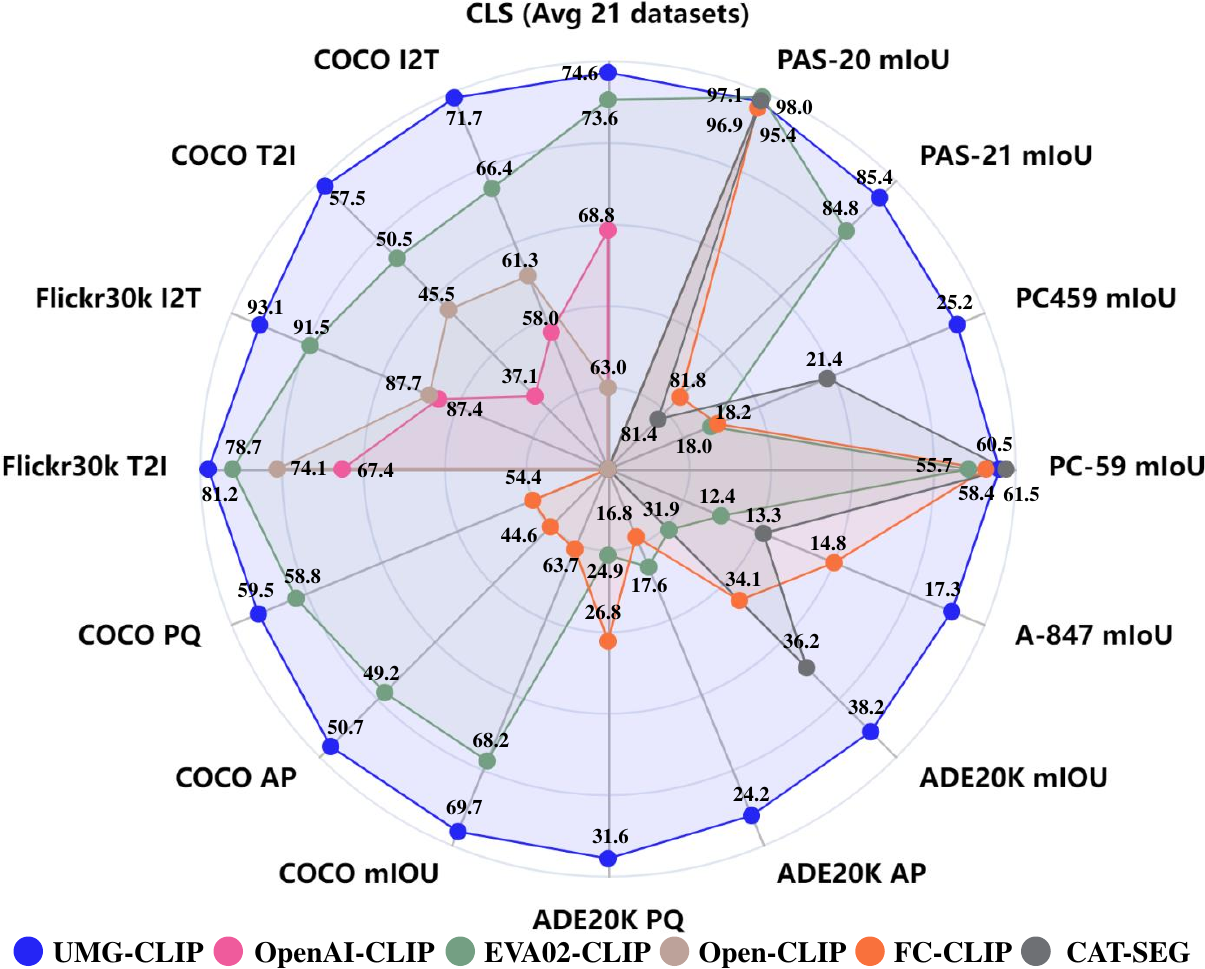}
    \caption{Compared to existing models, \modelname demonstrates outstanding performance across a wide range of tasks.}
    \label{fig:head_picture}
\end{figure} 


Recent years have witnessed remarkable progress towards developing \textit{Foundation Models} (FMs). Concurrent to Large Language models (LLMs) in natural language processing, vision-language FMs, represented by CLIP \cite{radford2021learning} and its variants \cite{li2023scaling,mu2022slip,ding2023open}, have gained significant prominence and become the mainstream for tasks that necessitate the joint understanding of both visual and textual modalities. Benefiting from the large-scale web-crawled data, these models \emph{de facto} manifest impressive performance gain for tasks that require global alignment, \emph{e.g.}, classification and retrieval. However, their performance tends to be comparatively suboptimal when confronted with dense tasks that require a more intricate understanding at the region or pixel level. This discrepancy arises from the primary focus of the learning paradigm of CLIPs, which emphasizes capturing the overall relationship between textual and images, rather than delving into fine-grained region-specific or pixel-level details. 


To overcome this limitation, considerable efforts have been devoted to enhancing the training granularity of CLIP. These endeavors primarily involve tailoring CLIP to specific tasks, such as object detection \cite{zhong2022regionclip,yao2022detclip,ma2023codet,wu2023cora}, segmentation \cite{ding2023open,wang2023samclip} and visual grounding \cite{chou2022semi}. However, due to the scarcity of high quality, fine-grained paired data, these studies often resort to implicitly mining fine-grained relations from image-text pairs. This procedure entails extracting relevant concepts from the textual information and establishing correspondences between these concepts and the corresponding regions/pixels within the image. However, these approaches introduce inherent noise in the annotations due to the implicit nature of the mining process. 


In this paper, we explore the integration of fine-grained localization ability into current CLIP models. The challenges mainly come from three aspects: the lack of high quality, fine-grained annotated dataset, together with an efficient training framework that can incorporate different levels of supervision, and furthermore, considering the task discrepancy, the low-cost adaptation paradigms for tackling various vision tasks. We address the above challenges in terms of both data and model perspectives. 
  
\begin{itemize}
    \item \textbf{Multi-granularity dataset generation.} 
    By leveraging the wide range of models available from the ML community,
    We are able to develop an automated annotation workflow that is capable of generating detailed annotations across different levels. Specifically, we utilize high-performance classification  \cite{zhang2023recognize,zhang2023gpt4roi,chen2023shikra}, detection \cite{cai2022bigdetection,wang2023v3det} and segmentation \cite{kirillov2023segment} models as annotators to automatically generate annotations at image, region, and pixel levels. Moreover, we assess the quality of the generated annotations and develop an automated filtering scheme to carefully select and integrate the final annotated data.  We annotate six public datasets \cite{sharma2018conceptual,changpinyo2021conceptual,ordonez2011im2text,krishna2017visual,kamath2021mdetr,deng2009imagenet}, and obtain \dataname, which contains approximately $41$M images, $389$M regions, spanning $11,741$ classes.

    \item \textbf{Multi-task pre-training paradigm.} We accordingly propose a unified multi-granularity learning framework, named \modelname, to endow CLIP with local perception ability. Specifically, in addition to conventional image-text matching, \modelname extracts regions from images and performs region-text matching to enhance its fine-grained understanding ability explicitly.  Moreover, \modelname incorporates these two levels of tag supervision to improve its category differentiation capabilities. To ensure efficiency, particularly when using high-resolution images for fine-grained pre-training, \modelname employs a cluster-based strategy \cite{li2023ailurus} and reduces $75\%$ of the visual tokens during pre-training.
    
    \item \textbf{Efficient downstream adaptation.} 
    Since \modelname has already been pre-trained with a generalist ability across tasks of various granularity, its downstream adaption burden can be significantly reduced. Therefore, we employ parameter-efficient tuning (PET) technology for adapting \modelname to different tasks, where the pre-trained backbone of \modelname is frozen, thereby preserving its existing knowledge and circumventing the resource-intensive process of extensive adjustments. Besides, \modelname incorporates learnable lightweight Convpass modules \cite{jie2022convolutional} into its backbone, as well as task-related decoders, to adapt to different downstream tasks.
\end{itemize}

Integrating the above ingredients produces a powerful vision framework. As shown in Fig.~\ref{fig:head_picture}, compared with current vision-language models, \modelname demonstrates exceptional performance across a wide range of downstream tasks with different levels of granularity. Notably, it achieves state-of-the-art (SOTA) results across multiple benchmarks, encompassing open-world recognition, retrieval, semantic segmentation, and panoptic segmentation.

\section{Related Work}
\label{sec:related}



\noindent \textbf{Models for Vision-language Understanding.} 
For vision understanding, there has been an increasing trend of shifting from a close-set setting to a more challenging, open-world understanding. Among them, CLIP \cite{radford2021learning} and ALIGN \cite{jia2021scaling} are pioneer works that trained with a large amount of image-text pairs. Following, FLIP \cite{li2023scaling} accelerates the training process by introducing masks. BLIP \cite{li2022blip} integrates understanding and generation tasks, enabling the model with captioning capabilities. Notably, EVA-Series\cite{fang2023eva, fang2023eva02, sun2023eva} adjust the model structure, scale up the training data and model size, and set up a series of state-of-the-art performance across different vision tasks. There are also works considering introducing local perception ability for CLIP. \cite{dong2023maskclip,yang2023attentive}  propose to align unmasked local patches with the global representation under semantic text supervision, rather than under specific region description supervision, which limits their ability to discriminate at the region level.  Furthermore, DetCLIP \cite{yao2022detclip} introduces a concept dictionary to enrich the concepts with their descriptions. CoDet \cite{ma2023codet} discovers the co-occurring objects across images and aligns them with the shared concept. MaskCLIP \cite{ding2023open} adds a mask generator to the frozen CLIP and calculates the similarity between mask and text embeddings to complete the zero-shot task. However, their annotation mining process inevitably introduces noise.


\begin{figure}[!t]
    \centering
    \includegraphics[width=0.85\linewidth]{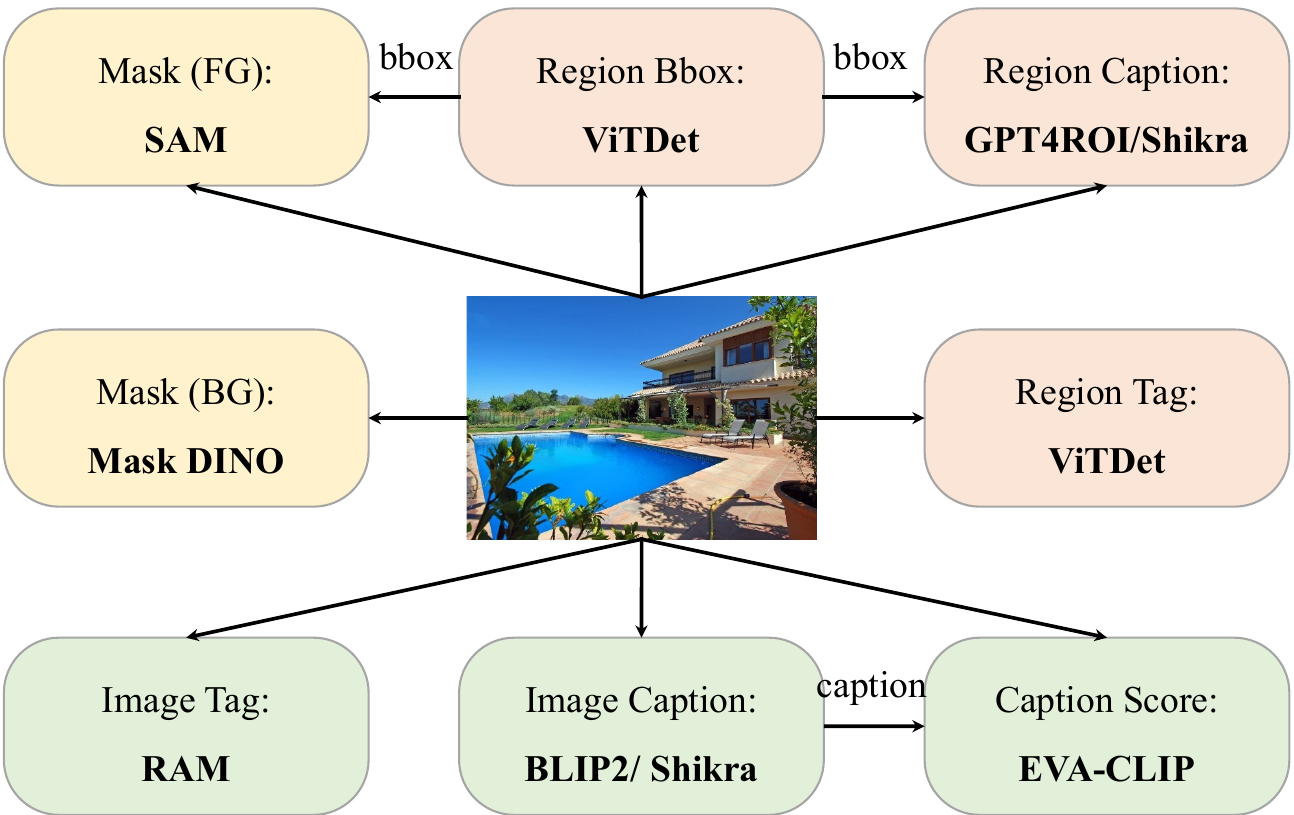}
    \caption{The automated annotation workflow, which generates captions/tags at image-level, region-level, and pixel-level.}
    \label{fig:annotation}
\end{figure}

\noindent \textbf{Dataset for Vision-language Understanding.}  
The vision-language datasets have been progressed from small scales, like SBU \cite{ordonez2011im2text}, Flickr30k \cite{young2014image}, and CC3M \cite{sharma2018conceptual}, to larger ones CC12M \cite{changpinyo2021conceptual}, LAION400M \cite{schuhmann2021laion}, COYO-700M \cite{coyo700m}, LAION5B \cite{schuhmann2022laion}. However, these datasets are limited with global image-text pairs. Recently, SA-1B \cite{kirillov2023segment} provides mask annotations of local regions, but lacks semantic information. While there have been efforts to describe local regions, such as Visual Genome \cite{krishna2017visual} and RefCOCO \cite{kazemzadeh2014referitgame}, these datasets remain relatively small in size. GLIP \cite{li2022grounded, zhang2022glipv2}, attempts to build word-region alignment data, but its word data is extracted from detection datasets and uses original image captions as noun phrases, which suffers certain deficiencies in terms of richness and quality. Recent work All-Seeing \cite{wang2023all} develops an automated annotation process that provides detailed text annotations for regions. However, its annotation granularity is limited to the region level. 

\begin{table*}[!t]
\centering
\scriptsize
\caption{Details of \dataname. We re-annotate six public datasets at multiple levels of granularity, resulting in a total of around $41$M images, $389$M regions, and spanning $11,741$ classes.}
\label{table:dataset}
\begin{tabular}{C{1.6cm} | C{1.4cm} C{1.3cm} C{1.2cm} C{1.3cm} C{1.9cm} C{1.3cm} | C{1.3cm}}
\Xhline{1pt}
Dataset      & CC3M \cite{sharma2018conceptual}   & CC12M \cite{changpinyo2021conceptual}  & SBU \cite{ordonez2011im2text}   & VG \cite{krishna2017visual}   & YFCC15M \cite{kamath2021mdetr} & IN21k \cite{deng2009imagenet}  & TOTAL   \\ \Xhline{0.5pt}
\#images     & 2.76M  & 8.60M  & 0.84M & 0.11M & 15.10M  & 14.00M & 41.41M  \\ 
\#regions    & 25.55M & 88.18M & 7.33M & 1.92M & 173.04M & 93.10M & 389.13M \\ 
\#categories & 2546   & 5292   & 2984  & 1372  & 5476    & 11287 & 11741   \\ \Xhline{1pt}
\end{tabular}
\end{table*}

\begin{figure*}[t!]
    \centering
    \includegraphics[width=\linewidth]{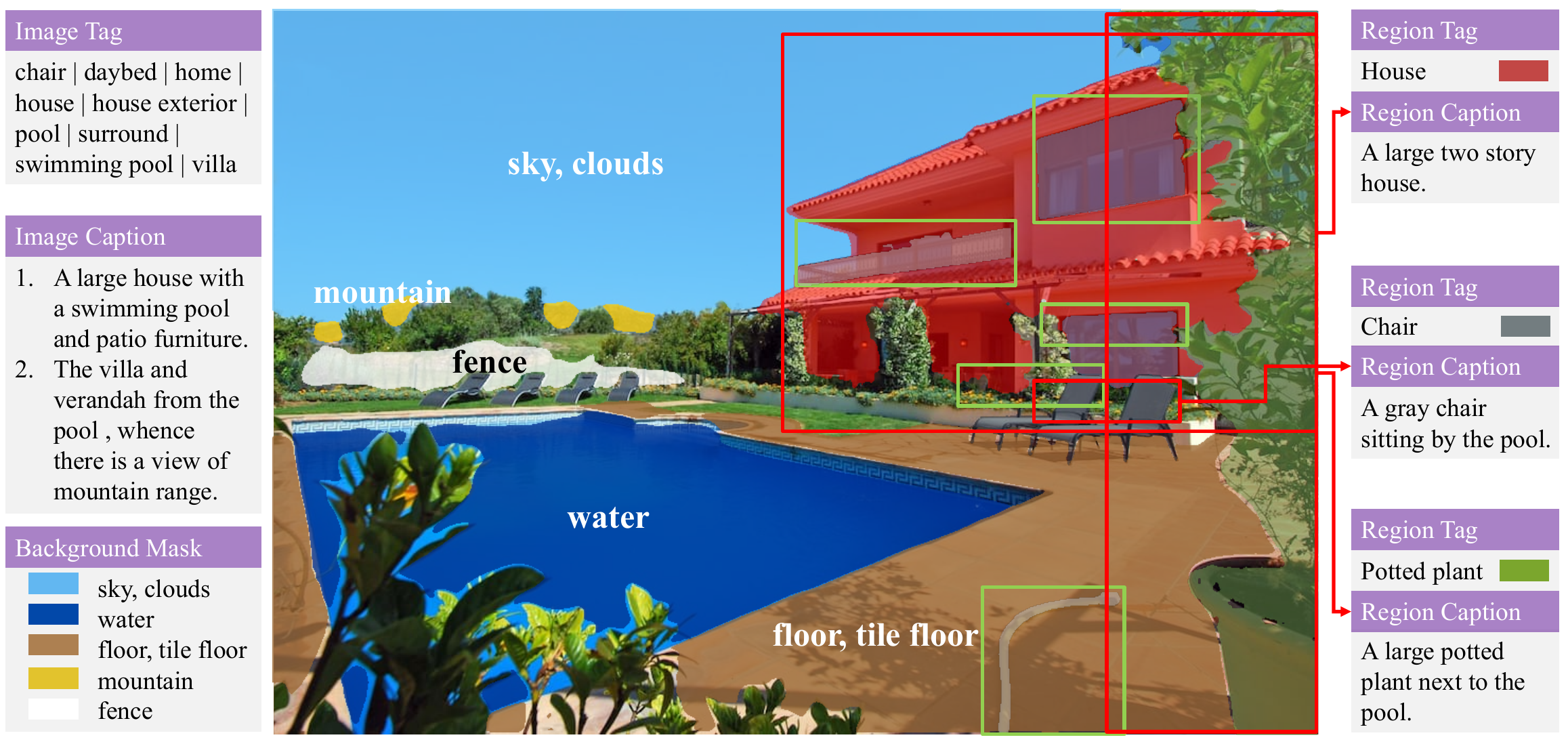}
    \caption{Visualization of  annotated image in \dataname. It includes tag and caption annotations at both image-level and pixel-level, as well as masks for both foreground and background.}
    \label{fig:dataset}
\end{figure*}

\section{\dataname Dataset}
\label{sec:dataset}


In this section, we elaborate on the details of our automated annotation workflow, as well as the labeled large-scale fine-grained dataset \dataname. 


\noindent \textbf{Data Annotation Workflow.} 
As shown in Fig.~\ref{fig:annotation}, we present an automated annotation workflow that facilitates comprehensive annotation, encompassing image-level, region-level, and pixel-level perspectives:

\begin{itemize}
    \item \noindent \textbf{Image-level.} For global images, we employ a high-performance tag model, RAM \cite{zhang2023recognize}, to annotate its tags. Similarly, we utilize BLIP2 \cite{li2023blip} and Shikra \cite{chen2023shikra} to provide the corresponding captions. Additionally, we utilize EVA-CLIP \cite{sun2023eva} to assign a score to each caption, which serves as an indicator for filtering high-quality captions. The obtained captions are also checked by expert scoring strategies, and details are included in the appendix.
    
    \item \noindent \textbf{Region-level.} For regional-level annotations, we first utilize two ViTDet \cite{li2022exploring} models, trained on BigDetection \cite{cai2022bigdetection} and V3Det \cite{wang2023v3det} separately, to generate candidate bounding boxes and their corresponding class labels. Subsequently, we filter out candidates with confidence scores below $0.3$ and apply Non-Maximum Suppression (NMS) to merge the remaining boxes. Based on the resulting boxes, we utilize the GPT4ROI \cite{zhang2023gpt4roi} and Shikra \cite{chen2023shikra} models to annotate their corresponding captions. 
     

    \item \noindent \textbf{Pixel-level.} For pixel-level annotations, we utilize two techniques: Mask DINO \cite{li2023mask} for generating background masks and SAM \cite{kirillov2023segment} for extracting foreground masks. The bounding-box prompts used for SAM are derived from our region-level annotations. In order to guarantee the quality of the generated foreground masks, we also assess their stability under bounding-box jittering and subsequently eliminate masks with low stability. More details are included in the appendix.
    
    
\end{itemize}



\begin{figure*}[t!]
    \centering
    \includegraphics[width=\linewidth]{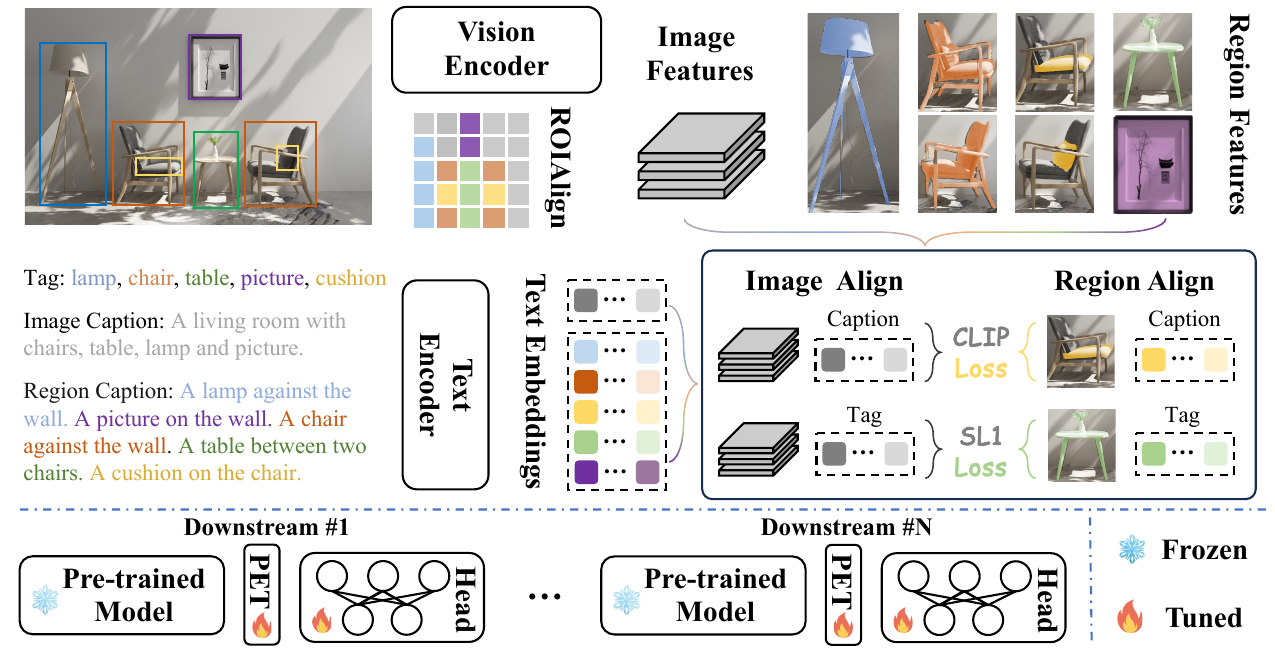}
    \caption{\modelname is a multi-granularity multi-task framework that aligns both image-level and region-level visual features with their corresponding tags and captions. This alignment empowers the model with generalist capabilities across multiple granularity, allowing it to efficiently adapt to various downstream tasks through PET.}
    \label{fig:framework}
\end{figure*}

\noindent \textbf{Dataset Details.}
We annotate a total of $41.41$M images from six public datasets, namely CC3M \cite{sharma2018conceptual}, CC12M \cite{changpinyo2021conceptual}, SBU \cite{ordonez2011im2text}, VG \cite{krishna2017visual}, YFCC15M \cite{kamath2021mdetr} and IN21K \cite{deng2009imagenet}. Our proposed dataset, named \dataname, encompassing annotations spans different levels of details. As illustrated in Tab.~\ref{table:dataset}, \dataname contains $389.13$M regions, each corresponding to one of $11,741$ distinct categories. Each region is associated with at least $4$ captions after eliminating duplicates, ensuring a diverse range of textual descriptions. We also provide global-level tags for each image, along with more than $7$ different captions that describe the image from various perspectives. These annotations are used for pre-training as described in Sec.~\ref{sec:stage1}. Additionally, we extend our efforts by providing pixel-level annotations for BigDet \cite{cai2022bigdetection}, resulting in the BigDet Panoptic dataset with $3.59$M images, $48.60$M regions, and $653$ classes. We leverage BigDet Panoptic for dense-level downstream adaptation, and more details are provided in Sec.~\ref{sec:stage2}. Fig.~\ref{fig:dataset} provides an example of the annotated data. More dataset statistics, including the category distribution, regional area distribution, and caption length, are included in the appendix. 



\section{Approach}

\label{sec:approach}

In this section, we introduce the pre-training methodology for the vision generalist model which possesses multi-granularity perception abilities, along with the techniques that efficiently adapt it to various downstream tasks.


\subsection{Unified Multi-granularity Learning Paradigm} 
\label{sec:stage1}

As described in Sec.~\ref{sec:dataset}, the \dataname dataset comprises annotations of varying granularity, including image-level tags $Y^I_{\text{tag}}$ and captions $Y^I_{\text{caption}}$, as well as region-level bounding boxes $Y^R_{\text{bbox}}$, tags $Y^R_{\text{tag}}$, and captions $Y^R_{\text{caption}}$.
Based on these fine-grained annotations, we train a robust foundation model termed \modelname with generalist capabilities.

\noindent \textbf{Framework Overview}
The overall framework of \modelname is shown in Fig.~\ref{fig:framework}. Specifically, given a mini-batch of input images $I$, we send it to the image encoder $E_I$ to obtain the visual embedding sequence $V_I = [v_{\text{cls}}, v_i] \in \mathbb{R}^{ b \times (n+1) \times d}$, where $v_{\text{cls}}$ and $v_i$ represents the class token and the visual tokens, respectively. $b$, $n$, and $d$ are the batch size, the visual token number, and the dimension, respectively. Note that we use EVA-CLIP \cite{sun2023eva} pretrained weights to initialize $E_I$ and the text encoder $E_T$. Then, we send $v_i$ and the bounding box annotations $Y^R_{\text{bbox}} \in \mathbb{R}^{  n_{\text{bbox}} \times 4}$,  where $n_{\text{bbox}}$ is the total region numbers in the input images, to the RoIAlign layer and produce the region embedding $v_r \in \mathbb{R}^{ n_{\text{bbox}} \times d}$. 
The above process can be denoted by: 
\begin{equation}
\begin{aligned}
v_{\text{cls}}, v_i &=  E_I(I), \\
v_r & = \operatorname{RoIAlign}( v_i, Y^R_{\text{bbox}}).
\end{aligned}
\end{equation}

Next, we pass the image-level annotations $Y^I_{\text{tag}}$ and $Y^I_{\text{caption}}$, as well as the region-level annotations $Y^R_{\text{tag}}$ and $Y^R_{\text{caption}}$, through the text encoder $E_T$ to obtain their respective embedding:
\begin{equation}
\begin{aligned}
t^I_{\text{tag}}, t^I_{\text{caption}} &=  E_T(Y^I_{\text{tag}}, Y^I_{\text{caption}}), \\
t^R_{\text{tag}}, t^R_{\text{caption}} &= E_T(Y^R_{\text{tag}}, Y^R_{\text{caption}}) ,
\end{aligned}
\end{equation}
where $t^I_{*} \in \mathbb{R}^{b \times d}$ and  $t^R_{*} \in \mathbb{R}^{n_{\text{bbox}} \times d}$. Finally, the total loss is calculated by considering the contribution of multi-granularity supervision:
\begin{equation}
\begin{aligned}
\mathcal{L}_{\text{image}} &= \mathcal{L}^I_{\text{tag}}(\hat{v}_{\text{cls}},\hat{t}^I_{\text{tag}})+ \mathcal{L}^I_{\text{caption}}(\hat{v}_{\text{cls}},\hat{t}^I_{\text{caption}}),\\
\mathcal{L}_{\text{region}} &= \mathcal{L}^R_{\text{tag}}(\hat{v}_{r},\hat{t}^R_{\text{tag}})+ \mathcal{L}^R_{\text{caption}}(\hat{v}_{r},\hat{t}^R_{\text{caption}}),\\
\mathcal{L}_{\text{total}} &= \alpha \mathcal{L}_{\text{image}} + \beta \mathcal{L}_{\text{region}},\\
\end{aligned}
\end{equation}
where $\hat{*}$ denotes the embedding after linear mapping. The contributions of the image-level loss and region-level loss are controlled by $\alpha$ and $\beta$ respectively, and we empirically set both of them to $1$. We utilize the smooth-L1 loss \cite{ren2015faster} for $\mathcal{L}^*_{\text{tag}}$ considering that different data may share the same or overlapping tags, while for $\mathcal{L}^*_{\text{caption}}$, we employ the visual-text contrastive loss \cite{radford2021learning}.

\noindent \textbf{Reducing Pretraining Memory.} 
Finer-grained pre-training requires higher pre-training resolution than the image-level one to ensure the pre-training quality, while increasing the resolution often results in heavier memory burdens. 
To deal with it, \modelname adopts a cluster-based strategy and reduces $75\%$ of the visual tokens during pre-training. Unlike FLIP \cite{li2023scaling} which randomly masks the visual tokens, we analyze that \modelname needs to preserve detailed spatial and semantic information for region-level learning. Thus, inspired by AliRus \cite{li2023ailurus}, we cluster the redundant tokens into representative tokens based on their intra-class similarity and inter-class distance and only propagate the representative tokens through subsequent transformer layers. At the end of the transformer, We unfold the tokens by assigning the value of the corresponding representative token to all merged positions to recover the original resolution for region-level learning.

\subsection{Task Adaptation with PET}
\label{sec:stage2}
Considering that \modelname has already been pre-trained with a generalist ability across tasks of various granularity, the burden of adapting it to different downstream tasks can be significantly reduced. We thus resort to parameter-efficient tuning (PET) technology to achieve this efficient downstream adaptation. Specifically, we freeze the pre-trained parameters of the \modelname backbone to retain its existing knowledge, while introducing learnable PET modules and task-specific decoders during the adaptation process to enhance its understanding of the specific downstream task. The PET modules are incorporated in parallel with the original multi-head self-attention (MHSA) and MLP layers of \modelname, and their computation can be represented by the following formula:
\begin{equation}
\begin{aligned}
X^{\prime} &= X + \operatorname{MHSA}(X) + s \cdot \operatorname{PET}_{\text{MHSA}}(X),\\
X^{\prime \prime} &= X^{\prime} + \operatorname{MLP}(X^{\prime}) + s \cdot \operatorname{PET}_{\text{MLP}}(X^{\prime}),\\
\end{aligned}
\end{equation}
where $s$ is the hyperparameter that controls the contribution of the PET module and is empirically set to $0.1$. $X$ is the input of the MHSA layer. We employ the Convpass \cite{jie2022convolutional} module with three convolutional layers ($1 \times 1$, $3 \times 3$, $1 \times 1$) and two GELU activations for both $\operatorname{PET}_{\text{MHSA}}$ and $\operatorname{PET}_{\text{MLP}}$.
To cater to different granularity requirements of the downstream, we have developed two distinct adapted models and further information can be found in Sec.~\ref{sec:implementation}.

\section{Experiments}
\label{sec:experiments}
\subsection{Implementation Details}
\label{sec:implementation}
\noindent \textbf{Pre-training Settings.}  
We follow EVA-CLIP \cite{sun2023eva} and maintain most of its training configurations during the pre-training phase. Specifically, we employ the LAMB optimizer with $\beta_1 = 0.9$, $\beta_2 = 0.98$, along with a weight decay of $0.05$. The learning rate for the visual encoder is set to $4e-4$ for \modelname-B/16 and $2e-4$ for other models. Similarly, the learning rate for the text encoder is set to $4e-5$ for \modelname-B/16 and $2e-5$ for other models. We use a $336 \times 336$ input resolution for all models, and the pre-training epochs are set to $6$ for \modelname-E/14 and $10$ for all other models. The pretraining lasts 2.37/4.29/5.54/6.12 days for UMG-CLIP-B/L/g/E, respectively, using 32 A800 GPUs. 
To ensure a smooth training process, we incorporate the warmup and learning rate decay schedule as outlined in \cite{sun2023eva}. Besides, we leverage the DeepSpeed optimization library \cite{rasley2020deepspeed} to conserve memory and expedite the training procedure. 

\begin{table*}[t!]
\centering
\footnotesize
\caption{Zero-shot classification performance on $21$ benchmarks. All results are evaluated at the resolution of $336$.}
\label{tab:classification}
\resizebox{\textwidth}{!}{
\renewcommand{\arraystretch}{1.5}
\begin{tabular}{c|ccccccccccccccccccccc|c}
\Xhline{2pt}
Model & \rotatebox{90}{Caltech101 \cite{fei2004learning}} & \rotatebox{90}{CIFAR10 \cite{krizhevsky2009learning}} & \rotatebox{90}{CIFAR100 \cite{krizhevsky2009learning}} & \rotatebox{90}{Country211 \cite{radford2021learning}} & \rotatebox{90}{DTD \cite{cimpoi2014describing}} & \rotatebox{90}{Eurosat \cite{helber2019eurosat}} & \rotatebox{90}{FER2013 \cite{goodfellow2013challenges}} & \rotatebox{90}{GTSRB \cite{stallkamp2012man}} & \rotatebox{90}{ImageNet-Ske. \cite{wang2019learning}} & \rotatebox{90}{ImageNet-Adv. \cite{hendrycks2021natural}} & \rotatebox{90}{ImageNet-Ren. \cite{hendrycks2021many}} & \rotatebox{90}{ImageNet-V2 \cite{recht2019imagenet}} & \rotatebox{90}{MNIST \cite{lecun1998gradient}} & \rotatebox{90}{ObjectNet \cite{barbu2019objectnet}} & \rotatebox{90}{PCam \cite{veeling2018rotation}} & \rotatebox{90}{Rendered SST2 \cite{radford2021learning}} & \rotatebox{90}{Resisc45 \cite{cheng2017remote}} & \rotatebox{90}{STL10 \cite{coates2011analysis}} & \rotatebox{90}{SUN397 \cite{xiao2010sun}} & \rotatebox{90}{SVHN \cite{netzer2011reading}} & \rotatebox{90}{VOC2007 \cite{everingham2015pascal}} & \rotatebox{90}{avg. top-1 acc.} \\ \Xhline{0.5pt}
OpenAI CLIP-B/16 \cite{radford2021learning} & 82.1 & 90.8 & 67.0 & \textbf{22.9} & 44.9 & 55.9 & 46.3 & \textbf{43.5} & 48.2 & 49.9 & 77.7 & 61.9 & \textbf{52.1} & 55.3 & 50.7 & \textbf{60.7} & 58.3 & 98.3 & 62.6 & 40.3 & 78.3 & 59.4 \\
Open CLIP-B/16 \cite{cherti2023reproducible} & 83.2 & 88.2 & 63.7 & 20.2 & \textbf{56.3} & 37.5 & \textbf{50.3} & 37.3 & 52.7 & 38.4 & 76.4 & 60.9 & 48.7 & 54.0 & 52.3 & 56.6 & 58.4 & 96.5 & 68.1 & \textbf{50.1} & 75.4 & 58.3 \\
EVA-02-CLIP-B/16 \cite{sun2023eva} & \textbf{85.9} & 96.8 & 82.3 & 22.6 & 52.2 & 63.0 & 49.7 & 41.1 & \textbf{55.9} & 61.9 & 80.7 & \textbf{67.7} & 41.1 & 63.2 & 53.1 & 54.2 & 59.6 & \textbf{99.4} & 68.3 & 26.1 & \textbf{78.5} & 62.1  \\
\rowcol \modelname-B/16 & 83.3 & \textbf{97.3} & \textbf{84.2} & 20.5 & 52.1 & \textbf{64.4} & 50.2 & 39.4 & 55.4 & \textbf{64.3} & \textbf{83.4} & 66.5 & 39.7 & \textbf{63.9} & \textbf{58.5} & 50.8 & \textbf{61.4} & \textbf{99.4} & \textbf{70.5} & 28.0 & \textbf{78.5} & \textbf{62.5} \\ \Xhline{0.5pt}
OpenAI CLIP-L/14 \cite{radford2021learning} & 83.4 & 94.9 & 74.4 & \textbf{34.4} & 55.7 & 61.4 & 49.2 & 52.5 & 61.0 & 77.5 & 89.0 & 70.9 &\textbf{79.0} & 72.0 & 60.8 & \textbf{70.6} & 63.8 & 99.4 & 66.6 & 50.3 & 78.1 & 68.8  \\
Open CLIP-L/14 \cite{cherti2023reproducible} & 84.1 & 93.8 & 71.4 & 24.1 & 61.4 & 52.6 & 46.9 & 45.0 & 57.0 & 46.4 & 81.7 & 64.2 & 65.2 & 58.2 & 57.1 & 55.7 & 61.8 & 97.8 & 70.6 & 50.5 & 76.7 & 63.0  \\
EVA-02-CLIP-L/14 \cite{sun2023eva} & \textbf{86.5} & \textbf{98.9} & \textbf{89.8} & 33.5 & \textbf{64.7} & \textbf{71.4} & 51.0 & 57.7 & \textbf{68.9} & \textbf{82.9} & \textbf{93.2} & \textbf{73.8} & 64.2 & \textbf{78.4} & 54.8 & 64.6& 69.0 & \textbf{99.7} & 72.5 & 45.8 & \textbf{82.7} & 71.6 \\
\rowcol \modelname-L/14 & 84.5 & 98.4 & 88.6 & 31.1 & 63.9 & 68.5 & \textbf{55.4} & \textbf{57.9} & 66.1 & 79.1 & \textbf{93.2} & 71.6 & 73.1 & 73.7 & \textbf{72.2} & 62.9 & \textbf{69.1} & 99.5 & \textbf{73.3} & \textbf{51.4} & 81.4 & \textbf{72.1} \\ \Xhline{0.5pt}
EVA-02-CLIP-E/14 \cite{sun2023eva} & 86.1 & \textbf{99.2} & \textbf{91.5} & 34.9 & 67.6 & 69.0 & \textbf{56.7} & 66.8 & \textbf{69.9} & 81.7 & 93.4 & \textbf{75.4} & 76.4 & 76.6 & 48.0 & 56.9 & \textbf{72.1} & 99.0 & 73.2 & 70.5 & \textbf{80.8} & 73.6  \\
\rowcol \modelname-E/14 & \textbf{86.2} & 98.6 & 90.2 & \textbf{35.0} & \textbf{67.9} & \textbf{69.3} & 56.4 & \textbf{71.0} & 67.9 & \textbf{83.3} & \textbf{94.5} & 72.7 & \textbf{84.4} & \textbf{76.8} & \textbf{50.8} & \textbf{61.4} & 71.0 & \textbf{99.6} & \textbf{75.1} & \textbf{76.0} & 79.3 & \textbf{74.6} \\ \Xhline{1pt}
\end{tabular}}
\end{table*}

\begin{table*}[t!]
\centering
\scriptsize
\caption{Zero-shot retrieval performance on COCO and Flickr30k. All results are evaluated at the resolution of $336$.}
\label{table:retrieval}
\setlength{\tabcolsep}{0.7mm}{
\begin{tabular}{c|cccccc|cccccc}
\Xhline{1pt}
\multirow{2}{*}{Model}                & \multicolumn{3}{c}{COCO I2T}                                                 & \multicolumn{3}{c|}{COCO T2I}                                                 & \multicolumn{3}{c}{Flickr30k I2T}                                            & \multicolumn{3}{c}{Flickr30k T2I}                                            \\ \cmidrule(r){2-4} \cmidrule(r){5-7} \cmidrule(r){8-10} \cmidrule(r){11-13}
 & R@1 & R@5 & R@10 & R@1 & R@5 & R@10 & R@1 & R@5 & R@10 & R@1 & R@5 & R@10 \\ \Xhline{0.5pt}  
OpenAI CLIP-B/16 \cite{radford2021learning} & 52.4 & 76.8 & 84.7 & 33.1 & 58.4 & 69.0 & 81.9 & 96.2 & 98.8 & 62.1 & 85.6 & 91.8 \\
Open CLIP-B/16 \cite{cherti2023reproducible} & 59.4 & 81.8 & 88.6 & 42.3 & 66.7 & 77.1 & 86.3 & 97.9 & 99.4 & 69.8 & 90.4 & 94.6 \\
EVA-02-CLIP-B/16 \cite{sun2023eva} & 58.7 & 80.7 & 88.2 & 42.2 & 66.9 & 76.3 & 85.7 & 96.7 & 98.9 & 71.2 & 91.0 & 94.7 \\
\rowcol \modelname-B/16 & \textbf{64.7} & \textbf{85.9} & \textbf{91.7} & \textbf{51.6} & \textbf{76.4} &\textbf{84.4} & \textbf{91.4} & \textbf{98.9} & \textbf{99.6} & \textbf{78.6} & \textbf{94.2} & \textbf{96.7}  \\ \Xhline{0.5pt}
OpenAI CLIP-L/14 \cite{radford2021learning} & 58.0 & 81.2 & 87.9 & 37.1 & 61.7 & 71.5 & 87.4 & 98.3 & 99.3 & 67.4 & 89.0 & 93.3 \\
Open CLIP-L/14 \cite{cherti2023reproducible} & 61.3 & 83.3 & 89.6 & 45.5 & 70.0 & 79.0 & 87.7 & 98.8 & 99.5 & 74.1 & 92.7 & 95.3 \\
EVA-02-CLIP-L/14 \cite{sun2023eva} & 64.2 & 85.2 & 90.8 & 47.9 & 71.7 & 80.0 & 89.2 & 98.9 & 99.6 & 77.9 & 94.2 & 96.8 \\
\rowcol \modelname-L/14  & \textbf{68.9} & \textbf{89.0} & \textbf{94.1} & \textbf{54.6} & \textbf{78.5} & \textbf{86.1} & \textbf{93.4} & \textbf{99.5} & \textbf{99.9} & \textbf{83.1} & \textbf{96.0} & \textbf{98.1} \\ \Xhline{0.5pt}
EVA-02-CLIP-E/14 \cite{sun2023eva} & 66.4 & 87.4 & 92.3 & 50.5 & 74.4 & 82.5 & 91.5 & 99.4 & 99.8 & 78.7 & 94.6 & 97.2 \\
\rowcol \modelname-E/14 & \textbf{71.7} & \textbf{89.8} & \textbf{94.3} & \textbf{57.5} & \textbf{80.4} & \textbf{87.3} & \textbf{93.1} & \textbf{99.7} & \textbf{99.9} & \textbf{81.2} & \textbf{95.7} & \textbf{97.7}  \\ \Xhline{1pt}
\end{tabular}}
\end{table*}

\noindent \textbf{Adaptation Settings.} 
We employ two distinct strategies for downstream adaptation based on the granularity of the tasks.
For global-level recognition tasks such as classification and retrieval, we utilize PET to adapt \modelname with the supervision of $\mathcal{L}_{\text{image}}$, leveraging the \dataname dataset. The learning rate is set to $1e-3$ using the LAMB optimizer. The input size remains unchanged from the pre-training stage. The number of adaptation epochs is set to $2$ for \modelname-E/14 and $5$ for all other models.
For dense-level recognition tasks like panoptic segmentation and semantic segmentation, we adapt \modelname using two different datasets: COCO Panoptic and our annotated BigDet Panoptic. We adopt the adaptation settings employed in FC-CLIP \cite{yu2023convolutions}, with the only differences being in the input resolution and adaptation epochs. Specifically, the learning rate is consistently set to $1e-4$ for all adaptations, employing the AdamW optimizer. For COCO Panoptic, we perform adaptation for $50$ epochs, utilizing an input resolution of $896 \times 896$. When working with BigDet Panoptic, we choose a reduced input resolution of $640 \times 640$ and adapt for $10$ epochs to minimize computational costs while maintaining satisfactory performance. 

\begin{table*}[t!]
\centering
\scriptsize
\caption{Open-vocabulary panoptic segmentation results on COCO and ADE20K.} 
\label{tab:panoptic}
\begin{tabular}
{c|cccccccccc}
\Xhline{1pt}
\multirow{2}{*}{Model}        & \multirow{2}{*}{Training dataset}   & \multicolumn{3}{c}{ADE20K zero-shot}          & \multicolumn{3}{c}{COCO zero-shot}                                                 & \multicolumn{3}{c}{COCO training} \\ \cmidrule(r){3-5} \cmidrule(r){6-8} \cmidrule(r){9-11}
                                &      & \multicolumn{1}{c}{PQ} & \multicolumn{1}{c}{AP} & \multicolumn{1}{c}{mIoU} & \multicolumn{1}{c}{PQ} & \multicolumn{1}{c}{AP} & \multicolumn{1}{c}{mIoU} & \multicolumn{1}{c}{PQ} & \multicolumn{1}{c}{AP} & \multicolumn{1}{c}{mIoU}\\ \Xhline{0.5pt}
MaskCLIP \cite{ding2023open}   &      COCO Panoptic            &           15.1              &            6.0             &              23.7            &             -            &             -            &                     -      &           -              &            -             &           -               \\
FreeSeg \cite{qin2023freeseg}   &     COCO Panoptic             &           16.3              &            6.5             &              24.6            &             -            &             -            &                     -      &           -              &            -             &           -               \\
ODISE \cite{xu2023open}  &    COCO Panoptic              &           12.6              &            14.4            &              29.9            &             -            &             -            &                     -      &           55.4              &            46.0             &           65.2               \\
FC-CLIP \cite{yu2023convolutions}   &     COCO Panoptic             &           26.8              &            16.8             &              34.1            &             -            &             -            &                     -      &           54.4              &            44.6             &           63.7               \\ \Xhline{0.5pt}
EVA-02-CLIP-B/16 \cite{sun2023eva}   &      COCO Panoptic            &           19.5              &            12.2             &              27.3            &             -            &             -            &                     -      &           54.8              &            44.1             &           65.9               \\
\rowcol \modelname-B/16   &      COCO Panoptic            &          \textbf{23.7}               &             \textbf{14.9}            &             \textbf{30.8}             &            -             &           -              &                    -       &          \textbf{54.8}               &            \textbf{44.6}             &             \textbf{66.0}             \\
EVA-02-CLIP-L/14 \cite{sun2023eva}   &      COCO Panoptic            &            23.5            &              15.9           &             30.3             &             -            &              -           &                    -       &           \textbf{58.1}              &             \textbf{48.7}            &            68.6              \\
\rowcol \modelname-L/14   &      COCO Panoptic            &           \textbf{25.2}              &             \textbf{17.4}            &              \textbf{34.4}            &              -           &              -           &                    -       &             58.0            &             48.5            &              \textbf{68.9}            \\
EVA-01-CLIP-g/14 \cite{sun2023eva}   &      COCO Panoptic            &             22.8            &           15.3              &            29.7              &             -            &              -           &                    -       &            57.6             &             47.6            &            67.5              \\
\rowcol \modelname-g/14   &      COCO Panoptic            &           \textbf{26.2}              &               \textbf{17.3}          &             \textbf{34.2}             &             -            &            -             &                     -      &             \textbf{57.7}            &           \textbf{48.2}              &             \textbf{67.9}             \\ 
EVA-02-CLIP-E/14 \cite{sun2023eva}   &      COCO Panoptic            &            24.9             &             17.6            &           31.9               &            -             &             -            &                    -       &            58.8             &           49.2              &             68.2             \\
\rowcol \modelname-E/14   &      COCO Panoptic            &             \textbf{26.2}            &              \textbf{19.1}           &            \textbf{34.9}              &           -              &             -            &                    -       &            \textbf{59.0}             &           \textbf{50.2}              &            \textbf{69.5}              \\ \Xhline{0.5pt}

\rowcol \modelname-B/16    &     Bigdet Panoptic            &            27.1             &          19.6               &          34.6                &           44.3              &          35.4               &               57.3            &            56.0             &              45.7           &            67.1              \\
\rowcol \modelname-L/14   &      Bigdet Panoptic            &          29.1               &          22.7               &         36.1                 &            46.0             &           37.7              &               59.1            &            58.9             &             49.7            &           68.9               \\
\rowcol \modelname-g/14   &      Bigdet Panoptic            &            30.0             &             23.5            &            36.0              &            46.4             &           39.8              &               58.8            &            58.5             &             49.6            &             68.5             \\
\rowcol \modelname-E/14   &      Bigdet Panoptic            &            \textbf{31.6}             &             \textbf{24.2}            &           \textbf{38.2}               &            \textbf{48.0}             &            \textbf{40.1}             &               \textbf{61.3}            &             \textbf{59.5}            &             \textbf{50.7}            &             \textbf{69.7}             \\ \Xhline{1pt}
\end{tabular}
\end{table*}

\begin{table*}[t]
\centering
\scriptsize
\caption{Open-vocabulary semantic segmentation performance in terms of mIoU on $6$ different benchmarks. }
\label{tab:segmentation}
\begin{tabular}{C{3.0cm}|C{2.4cm}C{1.0cm}C{1.0cm}C{1.0cm}C{1.0cm}C{1.0cm}C{1.0cm}}
\Xhline{1pt}
{Model} & {Training dataset} & {A-150} & {A-847} & {PC-59   } & {PC-459} & {PAS-21} & {PAS-20} \\ \Xhline{0.5pt} 
SPNet \cite{xian2019semantic} & Pascal VOC & - & - & 24.3 & - & 18.3 & -  \\ 
ZS3Net \cite{bucher2019zero} & Pascal VOC & - & - & 19.4  & - & 38.3 & -  \\
LSeg \cite{li2022language} & Pascal VOC & - & - & - & - & 47.4 & -  \\ \Xhline{0.5pt}
SimBaseline \cite{xu2022simple} & COCO Stuff & 15.3 & - & - & - & 74.5 & -  \\
ZegFormer \cite{ding2022decoupling} & COCO Stuff & 16.4 & - & - & - & 73.3 & -  \\
LSeg+ \cite{ghiasi2022scaling} & COCO Stuff & 18.0 & 3.8 & 46.5 & 7.8 & - & -  \\
OVSeg \cite{liang2023open} & COCO Stuff & 29.6 & 9.0 & 55.7 & 12.4 & - & 94.5  \\
SAN \cite{xu2023side} & COCO Stuff & 33.3 & 13.7 & 60.2 & 17.1 & - & 95.5  \\
CAT-Seg (L) \cite{cho2023cat} & COCO Stuff & 31.5 & 11.4 & 62.0 & 20.4 & 81.8 & 96.6 \\ 
CAT-Seg (G) \cite{cho2023cat} & COCO Stuff & 36.2 & 13.3 & 61.5 & 21.4 & 81.4 & 97.1 \\ 
\Xhline{0.5pt}
OpenSeg \cite{ghiasi2022scaling} & COCO Panoptic & 21.1 & 6.3 & 42.1 & 9.0 & - & -  \\
MaskCLIP \cite{ding2023open} & COCO Panoptic & 23.7 & 8.2 & 45.9 & 10.0 & - & -  \\
ODISE \cite{xu2023open} & COCO Panoptic & 29.9 & 11.1 & 57.3 & 14.5 & 84.6 & -  \\
FC-CLIP \cite{yu2023convolutions} & COCO Panoptic & 34.1 & 14.8 & 58.4 & 18.2 & 81.8 & 95.4  \\\Xhline{0.5pt}
EVA-02-CLIP-B/16 \cite{sun2023eva} & COCO Panoptic & 27.3 & 8.3 & 54.0 & 14.3 & 82.7 & 96.3  \\
\rowcol \modelname-B/16 & COCO Panoptic & \textbf{30.8} & \textbf{11.8} & \textbf{58.5} & \textbf{18.2} & \textbf{82.8} & \textbf{96.3}  \\
EVA-02-CLIP-L/14 \cite{sun2023eva} & COCO Panoptic & 30.3 & 11.2 & 57.1 & 17.6 & 84.5 & \textbf{97.9}  \\
\rowcol \modelname-L/14 & COCO Panoptic & \textbf{34.4} & \textbf{13.7} & \textbf{61.0} & \textbf{19.7} & \textbf{85.2} & \textbf{97.9}  \\
EVA-01-CLIP-g/14 \cite{sun2023eva} & COCO Panoptic & 29.7 & 9.3 & 54.8 & 15.9 & 84.5 & \textbf{97.6}  \\
\rowcol \modelname-g/14 & COCO Panoptic & \textbf{34.2} & \textbf{13.6} & \textbf{60.6} & \textbf{18.9} & \textbf{85.2} & 97.4  \\
EVA-02-CLIP-E/14 \cite{sun2023eva} & COCO Panoptic & 31.9 & 12.4 & 55.7 & 18.0 & 84.8 & \textbf{98.0}  \\
\rowcol \modelname-E/14 & COCO Panoptic & \textbf{34.9} & \textbf{13.9} & \textbf{60.2} & \textbf{19.2} & \textbf{85.3} & 97.3  \\ \Xhline{0.5pt}
\rowcol \modelname-B/16 & Bigdet Panoptic & 34.6 & 13.8 & 58.2 & 21.1 & 80.2 & 96.3  \\ 
\rowcol \modelname-L/14 & Bigdet Panoptic & 36.1 & 15.4 & 58.7 & 23.2 & 83.5 & 96.1  \\
\rowcol \modelname-g/14 & Bigdet Panoptic & 36.0 & 16.3 & 59.1 & 22.7 & 84.1 & 96.3  \\
\rowcol \modelname-E/14 & Bigdet Panoptic & \textbf{38.2} & \textbf{17.3} & \textbf{60.5} & \textbf{25.2} & \textbf{85.4} & \textbf{96.9}  \\ \Xhline{1pt}
\end{tabular}
\end{table*}

\begin{table}[t!]
\centering
\scriptsize
\caption{Parameters of the pre-trained \modelname models.}
\label{table:parameters}
\begin{tabular}{C{3.3cm}|C{2.8cm}C{2.8cm}C{2.8cm}}
\Xhline{1pt}
Model & Image parameters &Text parameters & Total parameters\\ \Xhline{0.5pt} 
\modelname-B/16 & 86M & 63M &149M\\
\modelname-L/14 & 304M& 124M&428M\\
\modelname-g/14 & 1.0B& 124M&1.1B\\
\modelname-E/14 & 4.4B& 354M &4.7B\\
\Xhline{1pt}
\end{tabular}
\end{table}

\begin{table*}[t]
\centering
\scriptsize
\caption{Ablations on different combinations of loss functions. Retrieval: average R@1 of I2T and T2I on COCO and Flickr30k; Classification: average top-1 accuracy over datasets shown in Tab.~\ref{tab:classification}; Segmentation: average mIoU over datasets shown in Tab.~\ref{tab:segmentation}. } 
\label{tab:losscombine}
\begin{tabular}{C{1.3cm}|C{1.3cm}|C{1.3cm}|C{1.3cm}|C{2.0cm}|C{2.0cm}|C{2.0cm}}
\Xhline{1pt}
\multicolumn{2}{c|}{Caption} & \multicolumn{2}{c|}{Tag}  & \multirow{2}{*}{Classification} & \multirow{2}{*}{Retrieval} & \multirow{2}{*}{Segmentation} \\ 
\cmidrule(r){1-2} \cmidrule(r){3-4} 
Image& Region & Image &Region & & &\\

\Xhline{0.5pt}
\Checkmark &  &  &  & \textbf{71.7} & \textbf{73.9} & 47.7 \\
\Checkmark & \Checkmark &  &  & 71.3 & 73.8 & 49.9 \\
\Checkmark & \Checkmark & \Checkmark &  & 71.3 & \textbf{73.9} & 50.4 \\
\Checkmark & \Checkmark & \Checkmark & \Checkmark & 71.0 & \textbf{73.9} & \textbf{50.7} \\
\Xhline{1pt}
\end{tabular}
\end{table*}

\begin{table}[t!]
\begin{minipage}[c]{.51\linewidth}
  \centering
  \caption{Loss type for $\mathcal{L}_{\text{tag}}$. We ablate the L1-loss and the contrastive loss.}
  \scriptsize
  \label{tab:a1}
  \begin{tabular}{l c c c}
    \toprule
    Type & Classification & Retrieval & Segmentation \\
    \hline
    L1&\textbf{71.0} &73.9 &\textbf{50.7}\\
    Contrastive&70.1 &\textbf{74.0} &50.6 \\
    \bottomrule
  \end{tabular}

\end{minipage}
\begin{minipage}[c]{.48\linewidth}
  \centering
  \caption{Loss hyperparameters $\alpha$ and $\beta$.}
  \scriptsize
  \label{tab:a3}
  \begin{tabular}{c c c c c}
    \toprule
    $\alpha$ & $\beta$ & Classification & Retrieval & Segmentation \\
    \hline
    1.0 & 0.5 & \textbf{71.0} & \textbf{73.9}& 49.9\\
    0.5 & 1.0 & 69.8 & 73.3& 50.6\\
    1.0 & 1.0 &\textbf{71.0} & \textbf{73.9}& \textbf{50.7}\\
    \bottomrule
  \end{tabular}
\end{minipage}
\end{table}

\begin{figure}[t]
\centering
\includegraphics[width=\linewidth]{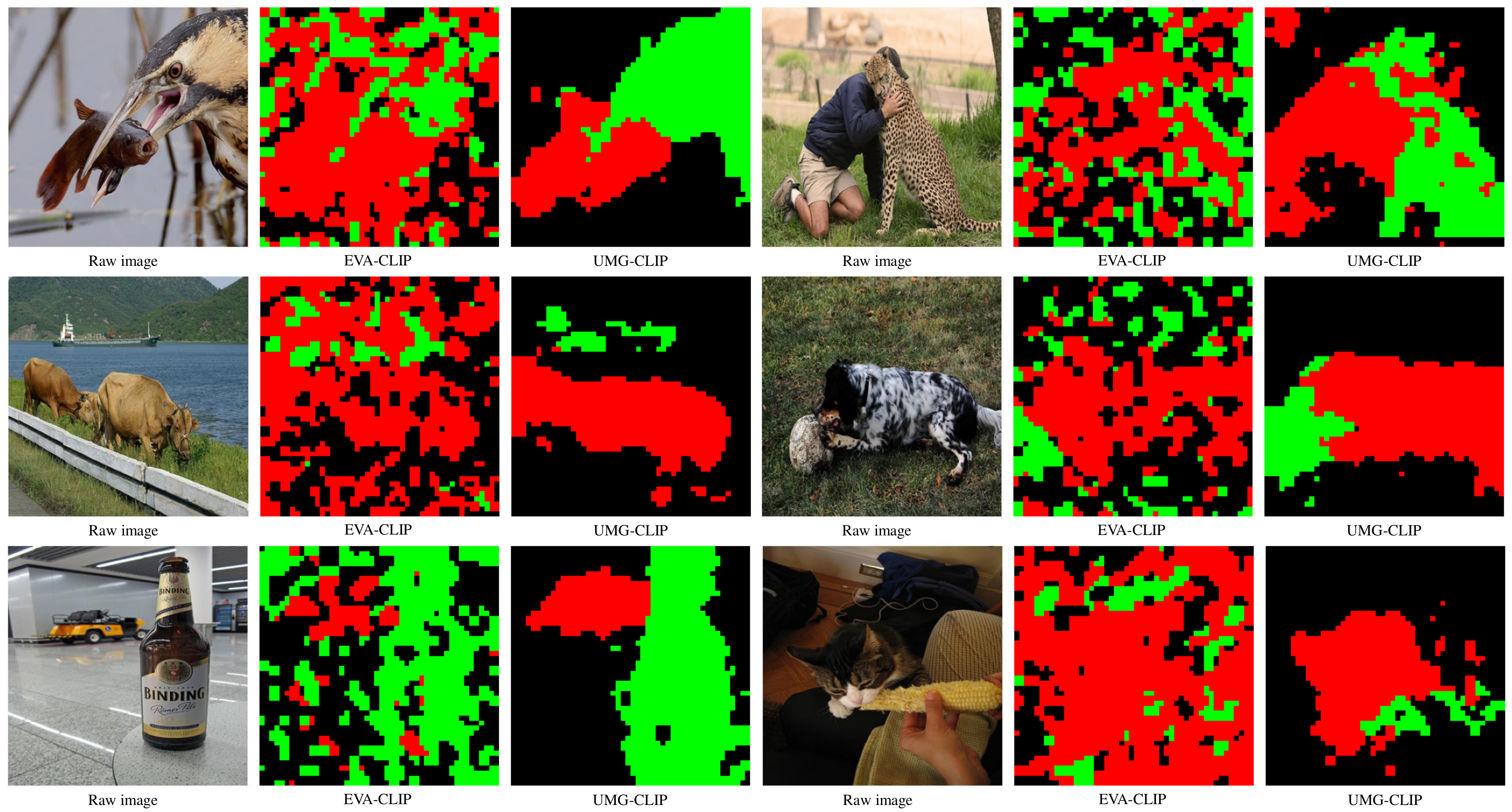}
    \caption{The alignment between up-sampled visual tokens and texts ("A photo of [tag]") using EVA-CLIP-L/14 and \modelname-L/14, respectively. "[tag]" corresponds to the class labels of the main objects in the images.}
    \label{fig:visalignment}
\end{figure}

\begin{table*}[t] 
\centering
\scriptsize
\caption{Ablations on PET and resolutions for dense-level downstream, reported with \modelname-L-14 trained on COCO Panoptic.}
\label{tab:pet}
\begin{tabular}
{C{1.8cm}C{1.4cm}C{1.3cm}C{1.3cm}C{1.3cm}C{1.3cm}C{1.3cm}C{1.3cm}}
\Xhline{1pt}
\multirow{2}{*}{Resolution} & \multirow{2}{*}{PET} & \multicolumn{3}{c}{ADE20K zero-shot} & \multicolumn{3}{c}{COCO training}  \\ \cmidrule(r){3-5} \cmidrule(r){6-8}
&  & \multicolumn{1}{c}{PQ} & \multicolumn{1}{c}{AP} & \multicolumn{1}{c}{mIoU} & \multicolumn{1}{c}{PQ} & \multicolumn{1}{c}{AP} & \multicolumn{1}{c}{mIoU}\\ \Xhline{0.5pt}
384  &  w/o & 23.9 & 14.3 & 32.6 & 52.6 & 39.5 & 66.7 \\
640  &  w/o & 23.6 & 14.7 & 32.7 & 54.2 & 42.3 & 65.3 \\
896  &  w/o & 23.2 & 14.3 & 32.5 & 53.2 & 42.4 & 65.2 \\\Xhline{1pt}
384  &  w   & 24.4 & 15.5 & 32.6 & 54.7 & 42.4 & 67.3 \\
640  &  w   & 24.8 & 16.9 & 33.4 & 56.5 & 46.0 & 68.0 \\
896  &  w   & \textbf{25.2} & \textbf{17.4} & \textbf{34.4} & \textbf{58.0} & \textbf{48.5} & \textbf{68.9}\\
\Xhline{1pt} 
\end{tabular}
\end{table*}

\subsection{Main Results}

\noindent \textbf{Zero-shot Classification.}  Tab.~\ref{tab:classification} presents the zero-shot classification results on $21$ different benchmarks. \modelname-L/14   achieves an average top-1 accuracy of $72.1\%$, which surpasses OpenAI CLIP-L/14 by $3.3\%$ and EVA-02-CLIP-L/14 by $0.5\%$. \modelname-E/14 further boosts the average performance to $74.6\%$, surpassing EVA-02-CLIP-E/14 by $1.0\%$. These results provide evidence that, even though \modelname introduces more attention to fine-grained discrimination, it remains highly competitive in global-level benchmarks.

\noindent \textbf{Zero-shot Retrieval.} Tab.~\ref{table:retrieval} presents the zero-shot retrieval results on COCO and Flickr30K datasets,  highlighting the significant performance advantage of \modelname over other methods. Notably, our approach achieves remarkable gains compared to EVA-02-CLIP-E/14, with R@1 improvements of $5.3\%$ and $7.0\%$ for COCO I2T and T2I, respectively. This substantial improvement can be attributed to the higher quality and increased diversity of text pairs in \dataname, which enhances the image-text understanding capabilities of \modelname.

\noindent \textbf{Open-vocabulary Panoptic Segmentation.} Tab.~\ref{tab:panoptic} inspects the panoptic segmentation results. Following FC-CLIP \cite{yu2023convolutions}, we first adapt \modelname using the COCO Panoptic dataset and subsequently evaluate its performance on both COCO and ADE20K. Remarkably, \modelname achieves comparable performance to EVA-02-CLIP on the \textit{seen} COCO dataset while significantly outperforming it on the \textit{zero-shot} ADE20K dataset. This superior zero-shot performance is also consistently maintained in semantic segmentation tasks, as demonstrated in Tab. \ref{tab:segmentation} and further discussed later. By leveraging our annotated Bigdet Panoptic dataset for adaptation,  \modelname further attains new state-of-the-art results across all benchmarks.

\noindent \textbf{Open-vocabulary Semantic Segmentation.} 
Tab.~\ref{tab:segmentation} presents the results of semantic segmentation on $6$ different benchmarks. Notably, our \modelname already demonstrates superior performance compared to both EVA-02-CLIP and FC-CLIP \cite{yu2023convolutions} when employing COCO Panoptic for adaptation. Moreover, when adapting with Bigdet Panoptic, the performance of \modelname exhibits further substantial improvements, particularly on demanding benchmarks such as A-150, A-847, and PC-459.

\noindent \textbf{Parameters Comparison.} Tab.~\ref{table:parameters} presents the parameters of the pre-trained \modelname models. Since the backbone of \modelname follows EVA-CLIP\cite{sun2023eva}, the parameters are comparable with it and other CLIP variants (\textit{e.g.}, OpenAI CLIP \cite{radford2021learning}, Open CLIP\cite{cherti2023reproducible}). However, as mentioned earlier, \modelname outperforms all of these works in terms of performance due to its more advanced pre-training paradigm. FC-CLIP \cite{yu2023convolutions} utilizes the ConvNeXt-L \cite{liu2022convnet} CLIP backbone from Open CLIP \cite{cherti2023reproducible} to better adapt to fine-grained downstream tasks. However, with the support of large-scale data, our \modelname-B/16 achieves superior performance compared to it while utilizing fewer image parameters ($86$M vs. $200$M). Similarly, CAT-Seg \cite{cho2023cat}, which uses both Swin-B \cite{liu2021Swin} and Open CLIP-G/14 \cite{cherti2023reproducible} backbones, employs more parameters ($1.6$B) but achieves worse performance compared to \modelname.

\subsection{Ablation Studies}
\noindent \textbf{Loss Combinations.} Tab.~\ref{tab:losscombine} presents the ablation results for different combinations of multi-task learning objectives. In this study and the studies for Tab.~\ref{tab:a1} and Tab.~\ref{tab:a3}, we exclude the PET modules and expedite the verification process by utilizing a subset of the data for both pre-training ($27$M data, with reduced epochs to $2$) and dense-level adaptation ($20\%$ COCO, with $384 \times 384$ input resolution). By incorporating region-level caption supervision, a notable enhancement in segmentation performance ($+2.2$ mIoU) is achieved compared to solely relying on image-level caption supervision. This improvement is accompanied by only a slight decline in classification ($-0.4\%$) and retrieval performance ($-0.1\%$). The addition of tag supervision leads to a further enhancement in segmentation performance, resulting in a mIoU of $50.7$, while still maintaining satisfactory levels of classification and retrieval performance. Note that we also attempt only to use the tag loss, but the performance collapses during the inference. We analyze that the form of $\mathcal{L}_{\text{tag}}$ indicates its primary function like a regularization term, which poses challenges in solely supporting downstream. 

\noindent \textbf{Loss type for $\mathcal{L}_{\text{tag}}$.}  As mentioned in Sec.~\ref{sec:stage1}, we employ L1-loss to avoid incorrect pushing away similar tags across different images and regions. Results in Tab.~\ref{tab:a1} confirm that the incorrect pushing away in contrastive loss adversely affects the classification performance, resulting in a $0.9\%$ performance drop.

\noindent \textbf{Loss hyper-parameters $\alpha$ and $\beta$.} Tab.~\ref{tab:a3} investigates the effect of adjusting the loss hyper-parameters $\alpha$ and $\beta$, proving that decreasing $\alpha$ hurts both classification and retrieval while decreasing $\beta$ degenerates fine-grained segmentation. Therefore, we do not further tune them and simply set them as $1.0$.

\noindent \textbf{PET and Resolution for Dense-level Downstream.}  Tab. \ref{tab:pet} presents a comprehensive analysis of the benefits of integrating PET modules for downstream adaptation, and the findings are encouraging. Models equipped with PET consistently outperform models without PET across the ADE20K and COCO datasets. Moreover, it has been noted in \cite{yu2023convolutions} that ViT-based models often struggle with generalization as the input size increases, while Tab. \ref{tab:pet} demonstrates that this issue can be mitigated through the implementation of PET, \textit{i.e.}, a continuous improvement in model performance is observed as the input resolution grows. 
These findings provide compelling evidence that the PET strategy we adopted can effectively enhance the downstream adaption ability of \modelname.

\noindent  \textbf{Correspondence between visual tokens and texts.} Fig.~\ref{fig:visalignment} shows a comparison of UMG-CLIP and its baseline EVA-CLIP in terms of correspondence between visual tokens and texts, based on images selected from the ImageNet-1K\cite{deng2009imagenet} validation set. 
Two main objects in the images are highlighted, with the red color representing one object and green representing the other. The visualization confirms that UMG-CLIP is capable of generating more detailed representations with improved correspondence, while EVA-CLIP \cite{sun2023eva} falls short in region-level discrimination. This intuitive observation demonstrates why UMG-CLIP outperforms EVA-CLIP in dense-level downstream tasks. 

\section{Conclusion}
\label{sec:conclusion}
This paper presents UMG-CLIP, a unified multi-granularity learning framework that endows the vanilla CLIP with precise local perception ability. UMG-CLIP tackles the challenges associated with both data and model training aspects. In terms of data, we propose an automated workflow capable of generating annotations at different levels by leveraging widespread models available from the ML community. This enables us to obtain annotations with varying granularities, enhancing the richness of the training data. Regarding model training, we put forward a multi-task learning strategy that simultaneously performs region-text matching as well as conventional image-text matching in both caption and tag dimensions, enabling UMG-CLIP to effectively learn and leverage the relations between regions and their corresponding textual descriptions. Equipped with the PET strategy, UMG-CLIP showcases exceptional performance advantages across various open-world understanding tasks. We hope that UMG-CLIP can serve as a valuable option for advancing vision-language foundation models, providing enhanced capabilities for various vision-language tasks and applications.


\noindent \textbf{Acknowledgement.} 
This work was supported in part by the National Natural Science Foundation of China under Grant 62125109, Grant 61931023, Grant 61932022, Grant 62371288, Grant 62320106003, Grant 62301299, Grant T2122024, Grant 62120106007.

\setcounter{section}{0}
\renewcommand\thesection{\Alph {section}}
\section{Dataset Statistics}

\begin{figure}[h!]
    \centering
    \includegraphics[width=\linewidth]{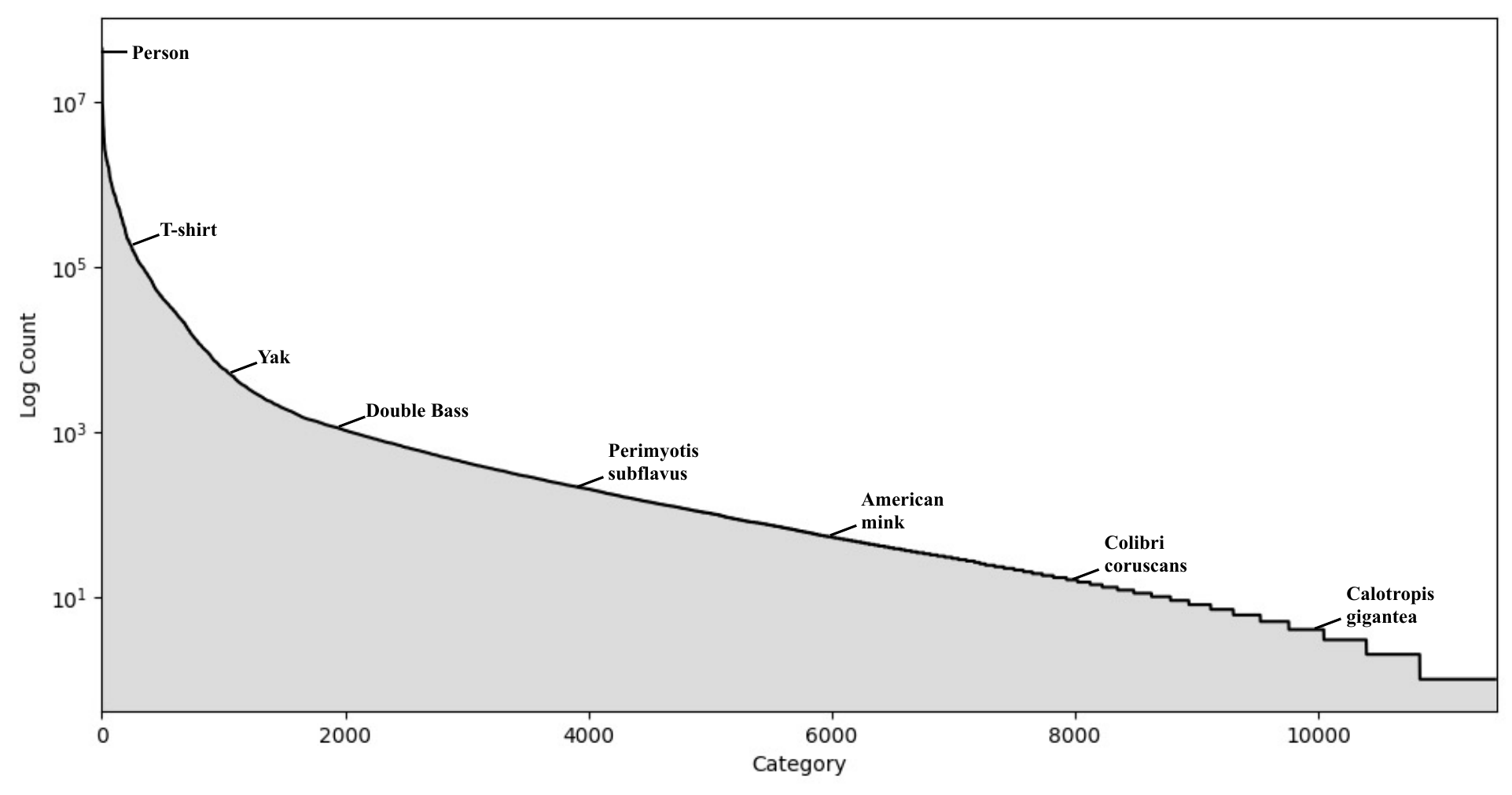} 
    \caption{Category Distribution of \dataname.}
    \label{fig:labelcount}
\end{figure}

\noindent \textbf{Category Distribution.}
Fig.~\ref{fig:labelcount} shows the statistics of category distribution of \dataname. In addition to common categories (\textit{e.g.}, people and T-shirt), \dataname also provides some more fine-grained category annotations (\textit{e.g.}, colibri coruscans and calotropis gigantea). The overall category exhibits a long-tail distribution.

\begin{table}[h!]
  \centering
    \caption{Regional Area Distribution of \dataname.}
    \setlength{\tabcolsep}{1.15cm}{
    \begin{tabular}{c|c|c}
      \Xhline{1pt}
      Region Type & Area Range & Proportion \\
      \Xhline{0.5pt}
      Tiny & $<20^2$ & 17.9\%\\
      Small & $20^2 \sim 40^2$ & 23.1\%\\
      Medium & $40^2 \sim 100^2$ & 29.1\% \\
      Large& $100^2 \sim 200^2$ &  16.0\%\\
      Huge & $>200^2$ & 13.8\%\\
      \Xhline{1pt}
    \end{tabular}}
    \label{tab:regioncount}
\end{table}

\begin{table}[t]
  \centering
    \caption{Average caption length of \dataname.}
    \setlength{\tabcolsep}{0.65cm}{
    \begin{tabular}{c|c|c|c|c}
      \Xhline{1pt}
      Type & Overall & BLIP2 & GPT4ROI & Shikra \\
      \Xhline{0.5pt}
      Image &7.50 &7.19 &-&10.96\\
      Region &5.85 & - & 5.59&6.33\\
      \Xhline{1pt}
    \end{tabular}}
    \label{tab:captionlen}
\end{table}

\begin{table}[t]
  \centering
    \caption{Caption quality evaluation results. The quality is evaluated by 4 experts among 400 samples. Percentages indicate the proportion of captions that are considered high quality to the total captions.}
    \setlength{\tabcolsep}{0.7cm}{
    \begin{tabular}{c|c|c|c}
      \Xhline{1pt}
      Raw Caption & Generated Caption & Both & None \\
      \Xhline{0.5pt}
      76.8\%& 47.3\% & 29.8\% & 5.8\%\\
      \Xhline{1pt}
    \end{tabular}}
    \label{tab:captionquality}
\end{table}

\begin{table}[t!]
  \centering
    \caption{Region Tag and bounding box quality evaluation results. The quality is evaluated by 4 experts among 100 samples with 680 regions. \Checkmark indicates that the corresponding annotation is considered accurate.}
    \setlength{\tabcolsep}{1.3cm}{
    \begin{tabular}{c|c|c}
      \Xhline{1pt}
      Tag & Bounding Box & Percentage \\
      \Xhline{0.5pt}
      \Checkmark & \Checkmark & 88.2\%\\
      \Checkmark & \XSolidBrush & 4.1\%\\
      \XSolidBrush & \Checkmark & 6.2\%\\
      \XSolidBrush & \XSolidBrush & 1.5\%\\
      \Xhline{1pt}
    \end{tabular}}
    \label{tab:tagbboxquality}
\end{table}

\noindent \textbf{Regional Area Distribution.} Tab.~\ref{tab:regioncount} examines the distribution of regional areas in \dataname. As \dataname primarily analyzes intricate scenes with multiple objects, it is observed that over $50\%$ of the annotated areas are small or medium-sized.

\noindent \textbf{Caption Length.} Tab.~\ref{tab:captionlen} inspects the average caption length of \dataname. \dataname uses an average of $7.50$ words to describe the overall information of an image, with the text annotations provided by Shikra being more specific ($10.96$ words). Region descriptions use relatively less text ($5.85$ words), as the information contained in a region is less than that of the entire image.

\begin{figure*}[t!]
    \centering
    \includegraphics[width=\linewidth]{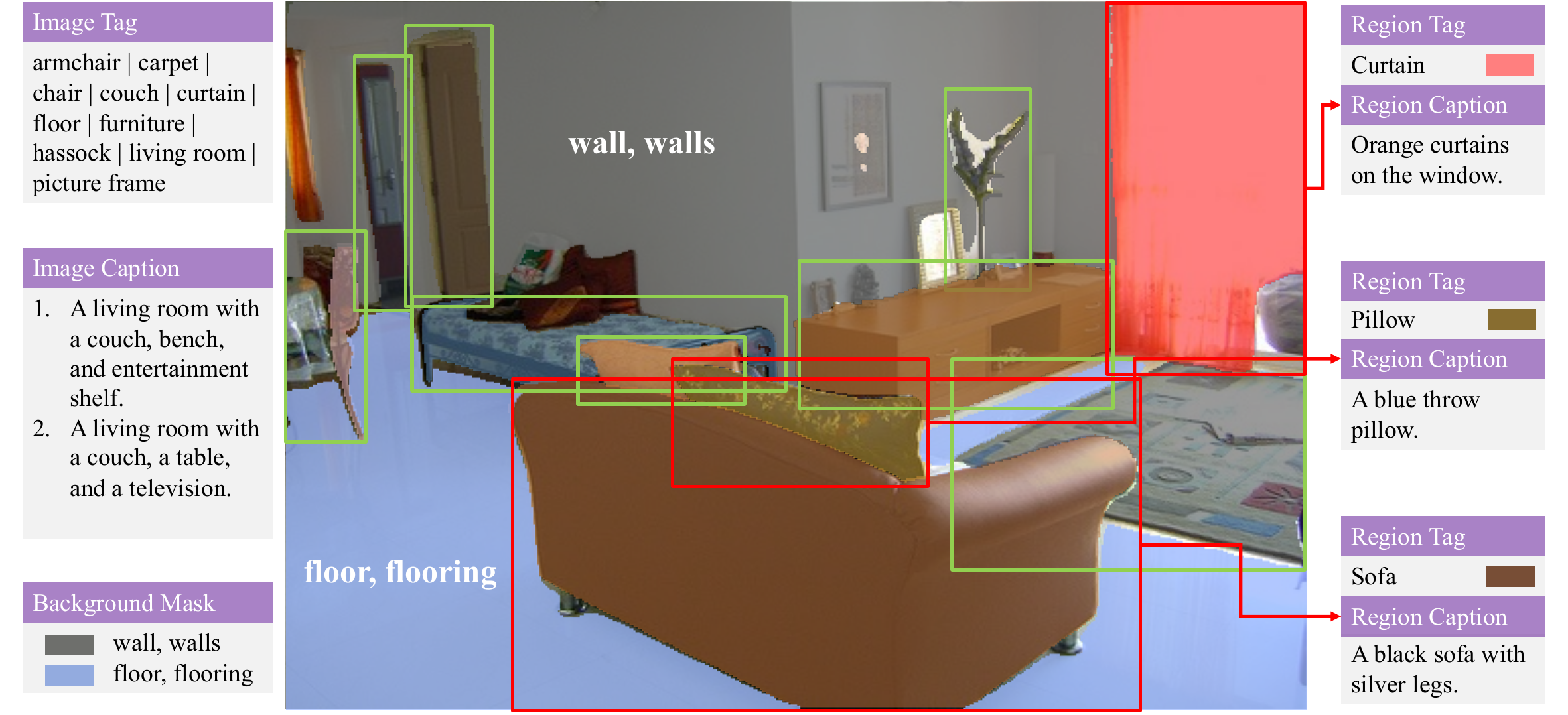}
    \centering
    \includegraphics[width=\linewidth]{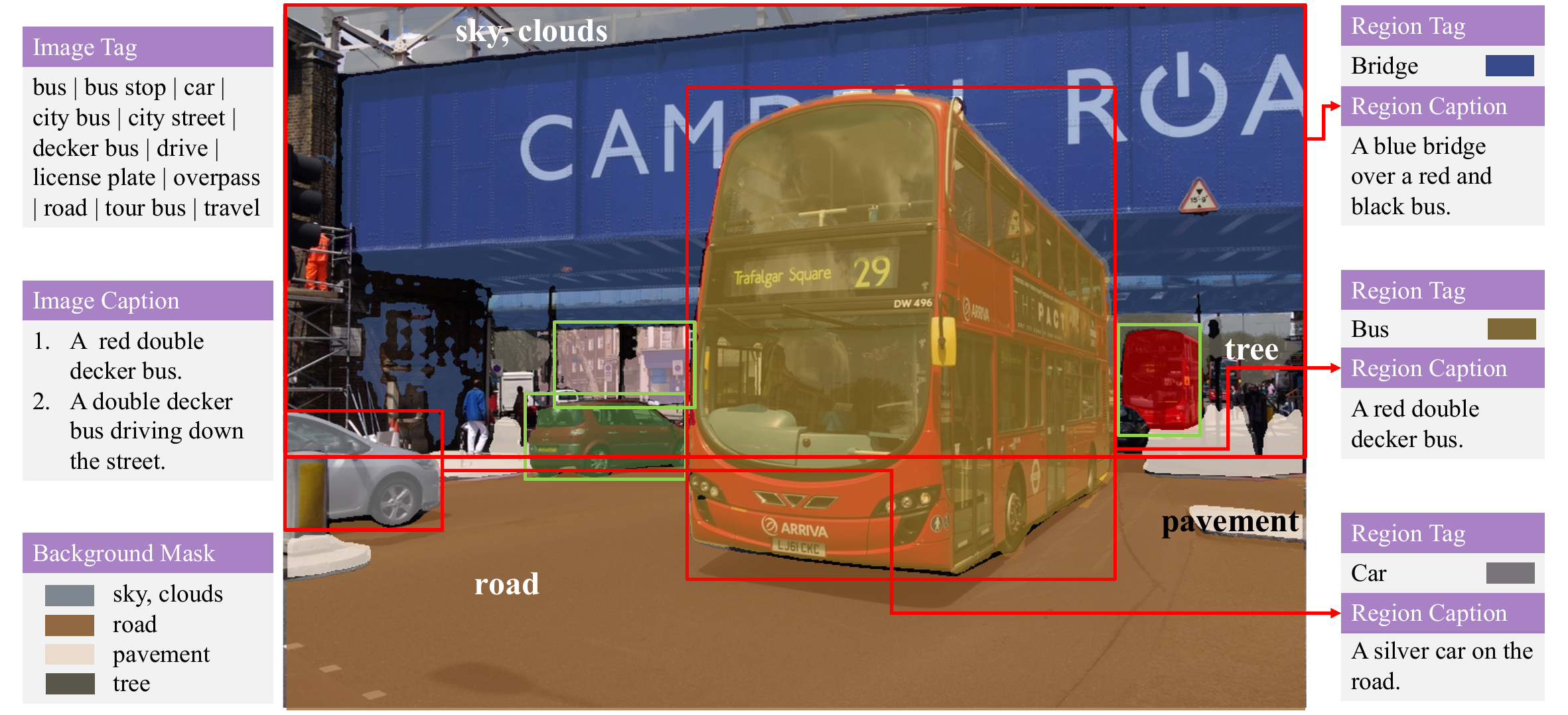}
    \caption{Visualizations of annotated examples in \dataname. }
    \label{fig:dataset2}
\end{figure*}

\section{Annotation Quality Evaluation}
\noindent \textbf{Caption Quality.} 
The caption quality is verified by human experts.  Specifically, four experts are involved in evaluating the quality of both the original captions provided by the raw datasets and the captions generated by our annotation system. A total of 400 samples are used in this assessment. The task of the experts is to determine which caption more accurately describes the content of the images.  Note that the experts also have the option to consider both captions as high quality or to choose none of them.
As shown in Tab.~\ref{tab:captionquality}, a higher percentage of the generated captions ($76.8\%$) are selected compared to the original captions ($47.3\%$), indicating that the quality of the generated captions is superior to that of the original captions.

\noindent \textbf{Region Tag and Bounding Box Quality.} Similarly, experts are also involved in the evaluation of the accuracy of tags and bounding box annotations for 680 regions across 100 samples. As shown in Tab.~\ref{tab:tagbboxquality}, $88.2\%$ of the regions are regarded to have both accurate tag and bounding box annotations, reflecting the high quality of the region-level annotations we have generated. Hence, we can confidently assert that our \dataname dataset is reliable.

\section{Bounding-box Jittering for Pixel-level Annotation}
To ensure the quality of the generated foreground masks, we evaluate their stability by subjecting them to bounding-box jittering and subsequently eliminate masks with low stability. This process involves slightly translating the original bounding box along the diagonal and using the translated bounding box to generate a new mask. Next, we calculate the pixel mIoU between the newly generated mask and the original mask. The stability score is then determined by the average mIoU obtained from multiple translations.

\section{Prompt Details}
We provide 9 and 274 distinct prompts for BLIP2 and Shikra, respectively, to generate image captions. Additionally, we offer 5 different prompts for Shikra and GPT4ROI to generate region captions. The full list of prompts will be included in our code page and here are some examples: "Provide a concise explanation of this photograph $\langle$image$\rangle$" (Shikra, image captions), "Please provide a detailed description of the $\langle$region1$\rangle$" (GPT4ROI, region captions).

\section{More Examples of Dataset Annotation}
In Fig.~\ref{fig:dataset2}, we present additional examples of dataset annotation to enhance the comprehensibility of our \dataname. These illustrations demonstrate the high quality of our annotations across diverse data samples.

\section{Limitation} 
We acknowledge that the current data annotation process can be cumbersome, involving the use of multiple additional models, and there is still room for improving the quality of annotations. 
In future work, we recognize the need for better synergy between the model training and data annotation to efficiently scale up the data collection procedure, as well as detailed analysis of the pseudo annotations.  


%
%
\bibliographystyle{splncs04}
\bibliography{main}
\end{document}